# Reversed Active Learning based Atrous DenseNet for Pathological Image Classification

Yuexiang Li, Xinpeng Xie, Linlin Shen* and Shaoxiong Liu

***Abstract*—Witnessed the development of deep learning in recent years, increasing number of researches try to adopt deep learning model for medical image analysis. However, the usage of deep learning networks for the pathological image analysis encounters several challenges, e.g. high resolution (gigapixel) of pathological images and lack of annotations of cancer areas. To address the challenges, we proposed a complete framework for the pathological image classification, which consists of a novel training strategy, namely reversed active learning (RAL), and an advanced network, namely atrous DenseNet (ADN). The proposed RAL can remove the mislabel patches in the training set. The refined training set can then be used to train widely used deep learning networks, e.g. VGG-16, ResNets, etc. A novel deep learning network, i.e. atrous DenseNet (ADN), is also proposed for the classification of pathological images. The proposed ADN achieves multi-scale feature extraction by integrating the atrous convolutions to the Dense Block. The proposed RAL and ADN have been evaluated on two pathological datasets, i.e. BACH and CCG. The experimental results demonstrate the excellent performance of the proposed ADN + RAL framework, i.e. the average patch-level ACAs of 94.10% and 92.05% on BACH and CCG validation sets were achieved.**

## I. Introduction

THE convolutional neural network (CNN) gains increasing attentions from the community since the AlexNet [1] won the ILSVRC 2012 competition. Currently, CNN has become one of the most popular classifiers in the area of computer vision. Due to its outstanding capacity, increasing number of researchers made their efforts to develop computer-aid diagnosis system using CNN. For example, Google brain proposed a multi-scale CNN model to aid breast cancer metastasis detection in lymph nodes [2]. However, several challenges arise while employing CNN for the classification of pathological images.

First, most of pathological images have a high resolution, i.e. gigapixel. Fig. 1 shows an example of thinprep cytological test (TCT) image for cervical carcinoma, with a resolution of 21163 x 16473.It is difficult for CNN to directly process the gigapixel images, due to the expensive computational cost.

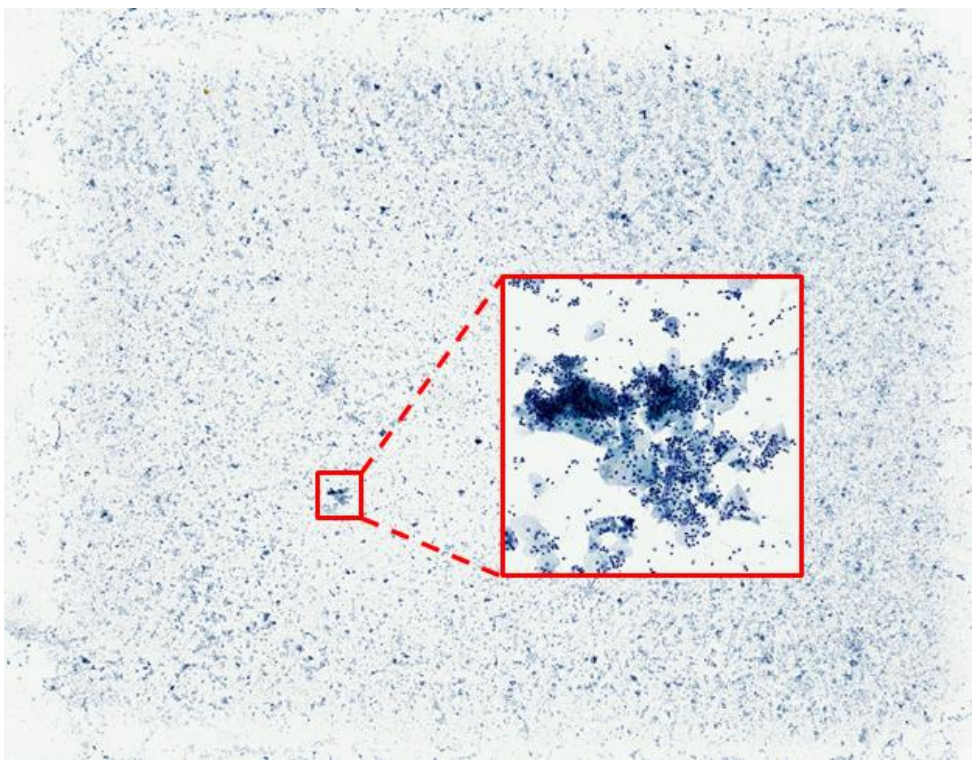

Fig. 1. The gigapixel TCT image for cervical carcinoma.

Second, some pathological datasets provide the images cropped from gigapixel samples. The image resolution is about 2000 x 1000 in this situation, which is possible for CNN to directly handle. However, the numbers of images contained in these dataset are usually small and insufficient to well train the CNNs. For example, the dataset used in 2108 grand challenge on breast cancer histology images (BACH) consists of 400 images with size 2048 x 1536 pixels. The dataset is divided to four categories. Therefore, each class only has 100 images. One of the solutions for the two problems is to train the CNN with patches cropped from either the gigapixel images, or the cropped images with size 2000 x 1000, and predict the label of a whole image based on the patch-level predictions. The cropping operation can decrease the size of training images and augment the volume of training set at the same time.

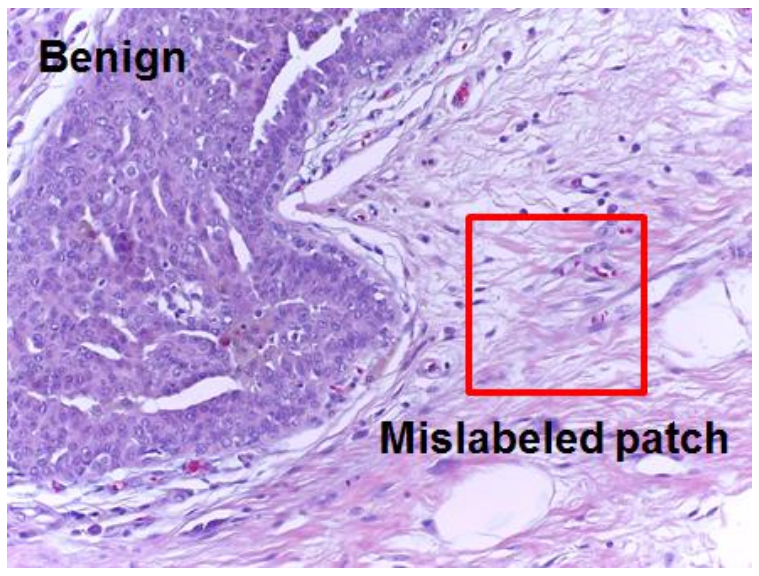

Fig. 2. An example of mislabeled patch from the BACH dataset. The normal patch is labeled as Benign.

The work was supported by Natural Science Foundation of China under grands no. 61672357, 61702339 and U1713214. Corresponding author: Prof. Linlin Shen

Yuexiang Li is with Youtu Lab, Tencent, Shenzhen, China. The work was completed when he worked as postdoctor in Computer Vision Institute, College of Computer Science and Software Engineering, Shenzhen University, Shenzhen, China. (Email: vicyxli@tencent.com)

Xinpeng Xie and Linlin Shen are with Computer Vision Institute, College of Computer Science and Software Engineering, Shenzhen University, Shenzhen, China. (Email: xiexinpeng2017@email.szu.edu.cn, llshen@szu.edu.cn)

Shaoxiong Liu is with the Sixth People's Hospital of Shenzhen, Shenzhen, China. (Email: liusx2008@sina.com)

As only the image-level ground truth is given in the classification task, the label of whole image is usually assigned to the corresponding generated patches. However, tumors may have a mixture of structure and texture properties [3] and there may be normal tissues around tumors. Hence, patch-level labels may be not consistent with the image-level label. Take the breast cancer histology image as an example (Fig. 2). The example patch contains normal tissue in Benign slice, but is labeled as Benign. These mislabeled patches may influence the subsequent network training and decrease the classification performance.

In this paper, we proposed a novel deep learning framework for the classification of pathological images. The main contributions can be summarized as follows:

1) A novel active learning strategy is proposed to remove the mislabeled patches from training set. When the traditional active learning is proposed to iteratively train a model with incremental manually labeled data, the proposed strategy, namely reversed active learning (RAL), can be seen as a reversed process of the traditional active learning.
2) An atrous DenseNet (ADN) is proposed for the classification of pathological images. We replace the common convolution of DenseNet to atrous convolution to achieve multi-scale feature extraction.
3) Experiments were conducted on two pathological datasets. The results demonstrate the outstanding classification performance of the proposed deep learning framework.

## II. Related Works

### A. Active Learning

Active learning (AL) aims to select appropriate unlabeled samples to be annotated by an oracle (e.g. a human annotator) and achieve high classification accuracy using as less labeled samples as possible [4]. The active learning approach first selects the most ambiguous/uncertain samples in the unlabeled pool for annotation and then retrain the machine learning model with the new labeled training set. Consequently, the size of labeled training set is incrementally augmented. The process is shown in Fig. 3.

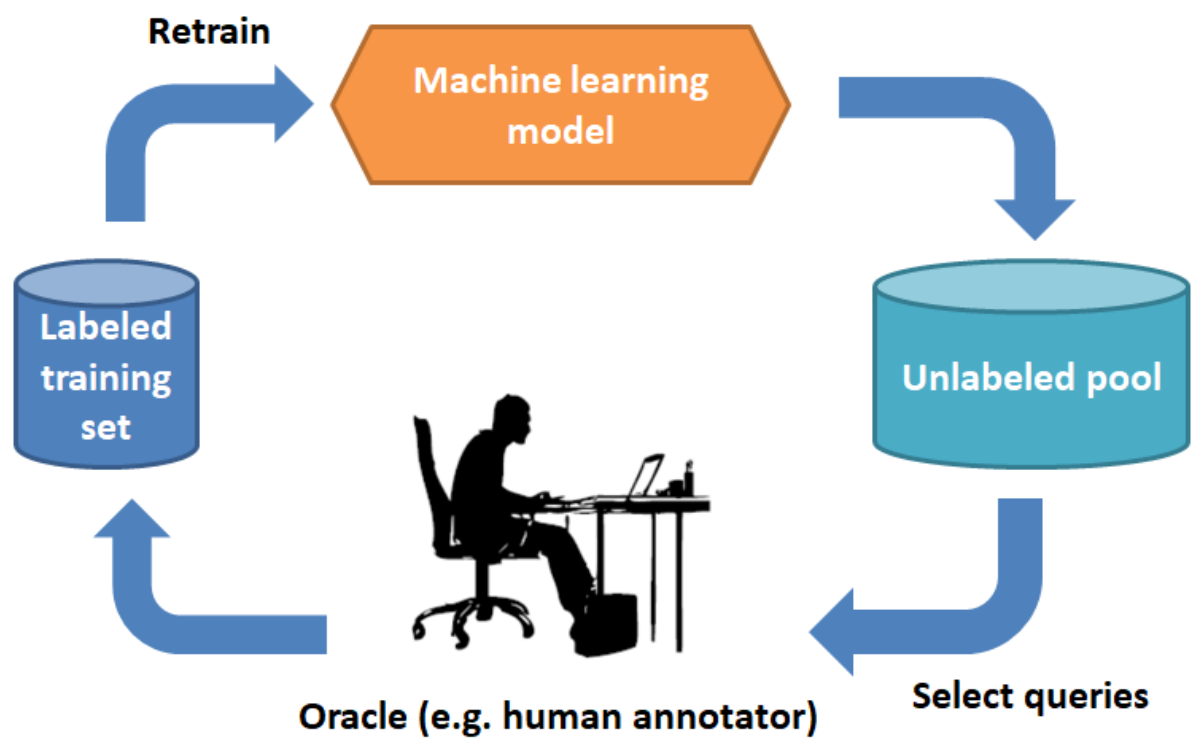

Fig. 3. The process of active learning. Ambiguous/uncertain samples are selected for oracle to incrementally augment the training set.

Wang proposed the first active learning approach for deep learning [4]. The approach uses three metric, i.e. least confidence, margin sampling and entropy, for data selection. Rahhal et al. suggested using entropy and Breaking-Ties (BT) as the confidence metric for the selection of electrocardiogram signals in the active learning process [5]. In recent research, researchers begin to employ active learning for medical image analysis. Yang proposed an active learning based framework, which consists of a stack of fully convolutional networks (FCN), to address the segmentation task of biomedical images [6]. The framework utilizes uncertainty and similarity information provided by FCN and formulates a generalized version of the maximum set cover problem to determine the most representative and uncertain areas for annotation. Zhou proposed a method called AIFT (active incremental fine-tuning) to naturally integrate active learning and transfer learning into a single framework [7]. The AIFT has been tested on three medical image datasets and achieved satisfactory results. Nan made the first try to employ active learning for the analysis of pathological images [8]. An improved active learning based framework, i.e. reiterative learning, was proposed to address the problem of partially labeled data for gastric cancer slides. The reiterative learning is an automatic approach without the interaction of oracle.

Although active learning is an extensively studied area, it is not appropriate for the task of patch-level pathological image classification. The aim of data selection for patch-level pathological image classification is to remove the mislabeled patches from the training set, which is different from the incrementally augmentation of training set of traditional active learning. To address the challenge, we proposed the reverse active learning (RAL) for patch-level data selection.

### B. Deep Learning based Pathological Image Analysis

The development of deep convolutional network was inspired by Krizhevsky who won the ILSVRC 2012 competition with the eight-layer AlexNet [1]. In the following competitions, growing numbers of participators developed novel networks, e.g. the VGG-nets [9] and the GoogLeNets [10]. The winner of ILSVRC 2015, He et al., proposed a much deeper convolutional network, ResNet [11]. The work addressed the training problem for deep convolutional nets and demonstrated that deeper networks indeed learn better features. Recently, the dense connected network (DenseNet) proposed by Huang outperformed the ResNet on various datasets [12].

With the development of deep learning technique, increasing number of deep learning based computer-aid diagnosis (CAD) models for pathological images have been proposed. Albarqouni developed a novel deep learning network, namely AggNet, for the mitosis detection in breast cancer histology images [13]. A completely data-driven model integrating numerous biological salient classifiers was proposed by Shah for invasive breast cancer prognosis [14]. Chen proposed a framework based on fully convolutional network for the segmentation of glands [15]. Li proposed an ultra-deep residual network for the segmentation and classification of human epithelial type 2 (HEp-2) specimen images [16]. In more recent researches, an end-to-end deep learning system was established by Liu to directly predict the H-Score for breast cancer tissue [17]. All the aforementioned algorithms cropped patches from pathological images to augment the training set and achieve satisfactory performance on specific tasks. However, we noticed that few of the presented researches involve the state-of-the-art network architecture, e.g. DenseNet, for the proposed CAD system, which leads to a margin for performance improvements. In this paper, we proposed a novel deep learning network, namely atrous DenseNet (ADN), for general analysis of pathological images. The proposed framework achieved the excellent classification performances on two types of pathological datasets, i.e. breast and cervical slices, which significantly outperform the benchmarking models.

## III. Reverse Active Learning

---

**Algorithm 1:** Reverse active learning

---

**Input:**
C: original training set C = $\{c_i\}$, i ∈ [1, n] {C has n patches}
$D_0$: augmented training set $D_0 = \{x_j^i\}$, j ∈ [1,8] {'rotation' & 'mirror' were adopted. $D_0$ has *8n* patches}
$M_0$: RN model pre-trained on $D_0$ {RN: a 6-layer CNN}
mx: counter {1 x n matrix }
**Output:**
$D_t$: refined training set at iteration t
$M_t$: fine-tuned RN model at iteration t
**Functions:**
p ← $P(x, M)$ {output of M}
$M_t \leftarrow F(D, M_{t-1})$ {fine-tune $M_t$ with D}
argmax(p): find the maximum value of vector p
zeros(mx): initialize all elements in matrix *mx* to zeros
**Initialize:**
t ← 1, zeros(mx)
**repeat**
    $D_t \leftarrow D_{t-1}$
    **for** each x ∈ $D_{t-1}$ **do**
        $p_j^i \leftarrow P(x_j^i, M_{t-1})$
        **if** argmax($p_j^i$) < 0.5 **then**
            remove $x_j^i$ from $D_t$
            mx(i) ← mx(i) + 1
        **end**
    **end**
    **if** ∀mx(i) ≥ 4 **then**
        remove $x_j^i$ from $D_t$
    **end**
    $M_t \leftarrow F(D_t, M_{t-1})$;
    t ← t + 1
**until** validation classification performance is satisfactory;

---

The patch-level datasets cropped from whole slice pathological images usually contain many mislabeled samples. To detect and remove the mislabeled patches, a reversed process of traditional active learning is proposed. As the existing deep-layer networks are easy to overfit to the training set, they are likely to remember the patches and fail to remove the mislabeled ones. To alleviate the overfitting problem, a simple 6-layer CNN, namely RefineNet (RN), was adopted for RAL, whose architecture is presented in *Appendix*. Let M represents the RN model in CAD system and D represents the training set with m patches (x), the process of reversed active learning (RAL) is illustrated in Alg. 1.

The RN model was first trained on original patch training set. Then, the trained model was adopted to make predictions on the training set. The patches, whose maximum confidence is lower than 0.5, were removed. Furthermore, since each patch was augmented to eight patches with data augmentation ('rotation' and 'mirror'), if more than four of the augmented patches were removed, the remaining patches were removed from the training set, either. A fixed validation set annotated by pathologists was adopted to evaluate the performance of finetuned model. Using the RAL, the amount of mislabeled patches gradually decreases. As a result, the performance of RN model on validation set was gradually improved. The RAL stops until the validation classification accuracy is satisfactory/stops increasing. The training dataset filtered by RAL can be seen as correctly annotated, and can be used to train deeper networks like ResNet, DenseNet, etc.

## IV. Atrous DenseNet (ADN)

### A. *Atrous Convolution*

Atrous convolution (or dilated convolution) was employed for deep learning based semantic segmentation approaches in many recent researches [18]. Compared to common convolution layer, the convolutional kernels in atrous convolution layer have 'holes' between their parameters, which enlarge the receptive filed and preserve the number of trainable parameters. The size of 'holes' inserted to the parameters is decided according to the dilation rate (γ). As shown in Fig. 4, a smaller dilation rate results in a more compact kernel, i.e. the common convolution is with dilation rate = 1, while a larger dilation rate produces an expanded kernel. A kernel with large dilation rate can capture more context information from the feature maps of previous layer.

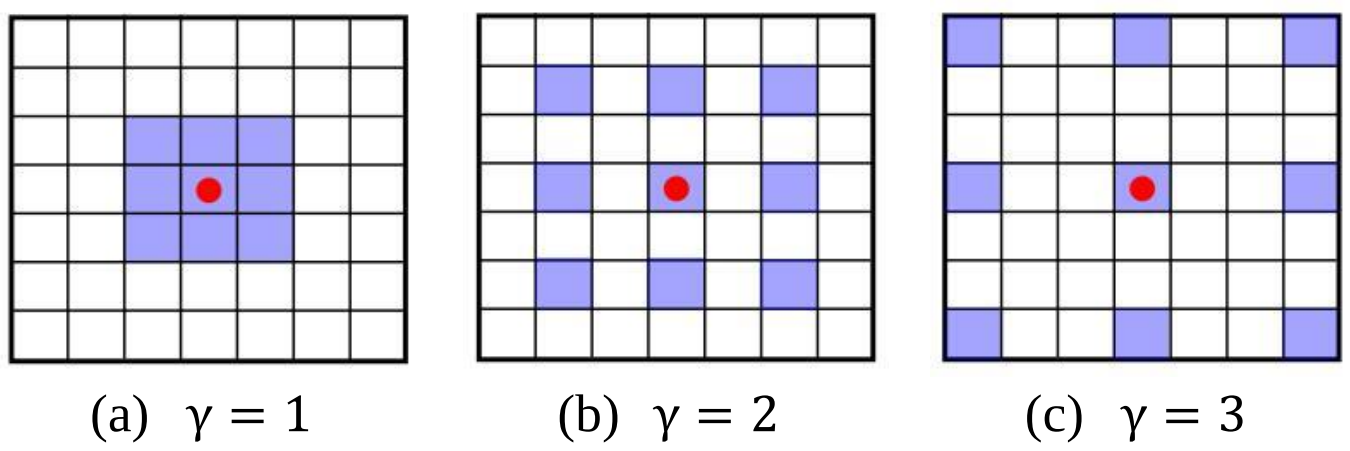

Fig. 4. Example of atrous convolutions with different dilation rates. The purple squares represent the positions of kernel parameters.

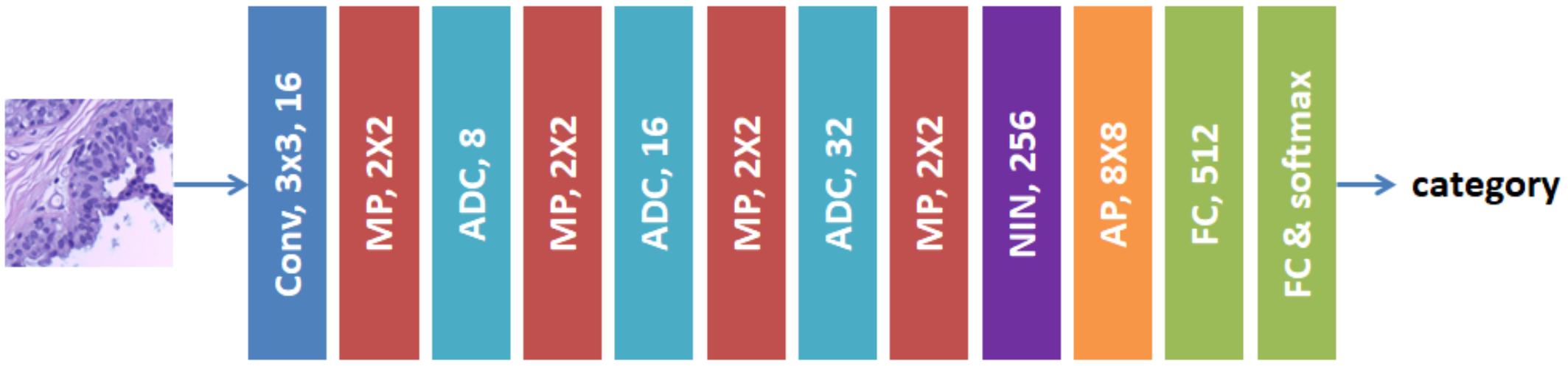

Fig. 5. Network architecture of proposed atrous DenseNet (ADN). Two modules, i.e. Atrous Dense Connection (ADC) and Network in Network (NIN) are involved in the ADN. The blue, red, orange and green rectangles represent the convolution, max pooling, average pooling and fully connected layers, respectively.

### B. Network Architecture

Using the atrous convolutions, multi-scale features can be extracted without extra computation cost, compared to that of approaches using common convolution kernels [10]. Taking the advantages into account, we proposed a novel deep learning architecture, i.e. atrous DenseNet, for pathological image classification. The network architecture is presented in Fig. 5. The blue, red, orange and green rectangles represent the convolutional layer, max pooling layer, average pooling layer and fully-connected layer, respectively. The proposed deep learning network has different architectures for shallow layers, i.e. Atrous Dense Connection (ADC), and deep layers i.e. Network in Network module (NIN) [19], for better feature extraction. PReLU is used as the non-linear activation function. The network optimization is supervised by the softmax loss (L), as defined in (1).

$$L = \frac{1}{N}\sum_i L_i = \frac{1}{N}\sum_i -\log(\frac{e^{f_{y_i}}}{\sum_j e^{f_j}}) \quad (1)$$

where $f_j$ denotes the $j$-th element ($j \in [1, K]$, $K$ is the number of classes) of vector of class scores $\boldsymbol{f}$, $y_i$ is the label of $i$-th input feature and $N$ is the number of training data.

### C. Atrous Dense Connection (ADC)

The Atrous Dense Connection (ADC) is inspired by the state-of-the-art network architecture, i.e. DenseNet [12]. Two primary developments were made to transform the original DenseNet block to our ADC: *a)* The ADC replaces the common convolutions in original DenseNet to atrous convolutions. *b)* A wide, instead of deep, DenseNet is built.

We first give the definition of original dense block and then introduce the network developments in details. The traditional Dense block is shown in Fig. 6. Let $H_l(.)$ be a composite function of operations such as convolution, rectified linear units (ReLU), the output of the $l^{th}$ layer ($x_l$) for a single image $x_0$ can be written as:

$$x_l = H_l([x_0, x_1, \dots, x_{l-1}]) \quad (2)$$

where $[x_0, x_1, \dots, x_{l-1}]$ refers to the concatenation of the feature maps produced by layers 0, … , $l$ – 1.

If each function $H_l(.)$ produces $k$ feature maps, the l$^{th}$ layer consequently has $k_0 + k \times (l-1)$ input feature maps, where $k_0$ is the number of channels in the input layer. $k$ is called growth rate of DenseNet block.

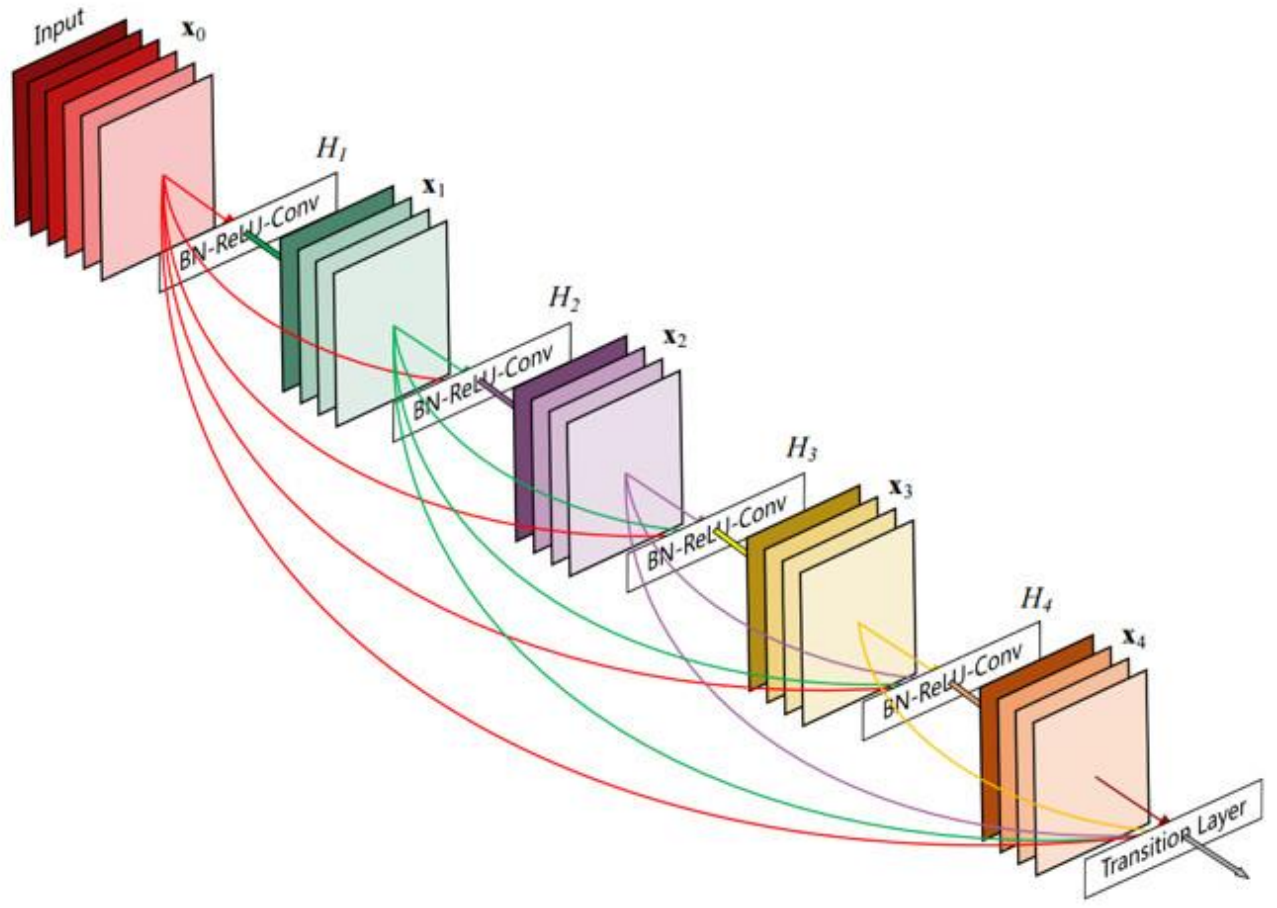

Fig. 6. Typical dense block adopted in DenseNet. The block involves four BN-ReLU-Conv modules. The color squares are the feature maps generated by different stages. The Conv is common convolution. The grow rate is 4.

#### 1) Atrous Convolution Replacement

The original dense block achieves multi-scale feature extraction by stacking 3 x 3 convolutions. As the atrous convolution has larger receptive field compared to the common convolution, the proposed Atrous Dense Connection block adopts atrous convolutions to achieve multi-scale feature extraction. As shown in Fig. 7. Atrous convolutions with two dilation rates, i.e. 2 and 3, are involved in the proposed ADC block. The common 3 x 3 convolution is placed following each atrous convolution to fuse the extracted feature maps and refine the semantic information.

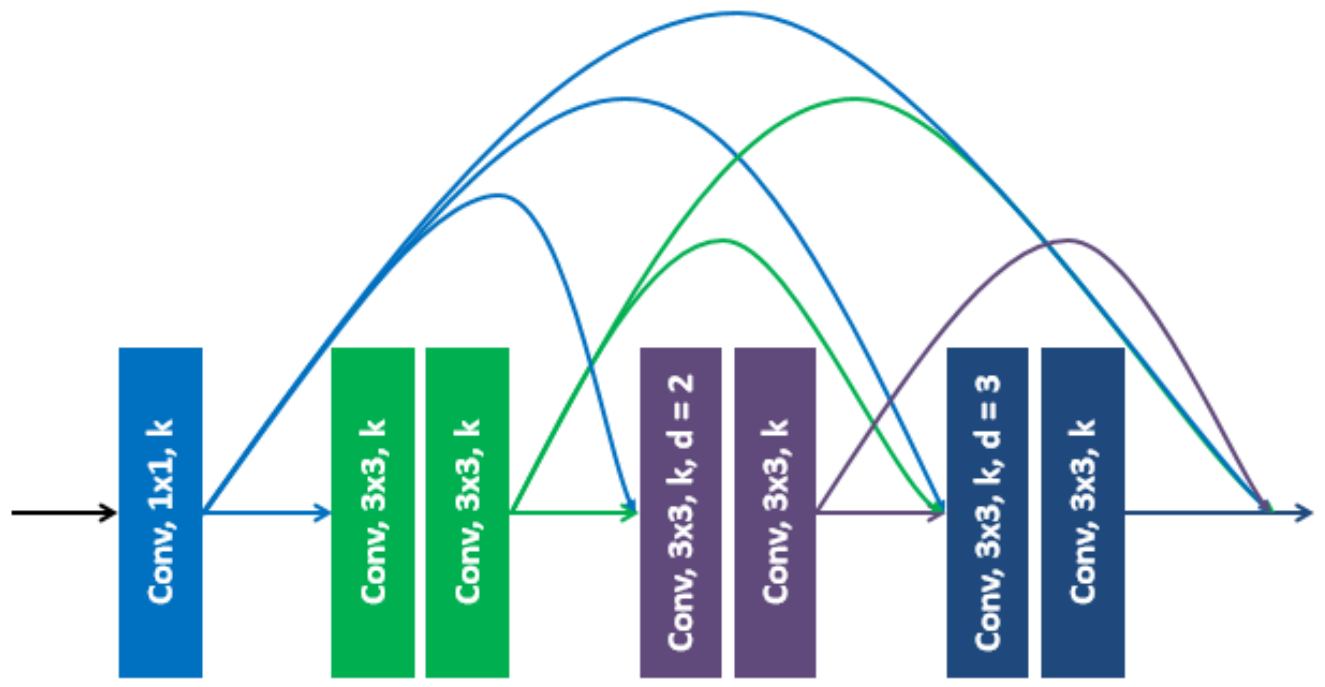

Fig. 7. Network architecture of proposed Atrous Dense Connection (ADC). Convolutions with different dilation rates are adopted for multi-scale feature extraction. Color connections refer to the feature maps produced by the corresponding convolution layers. The feature maps from different convolution layers are concatenated to form a multi-scale feature.

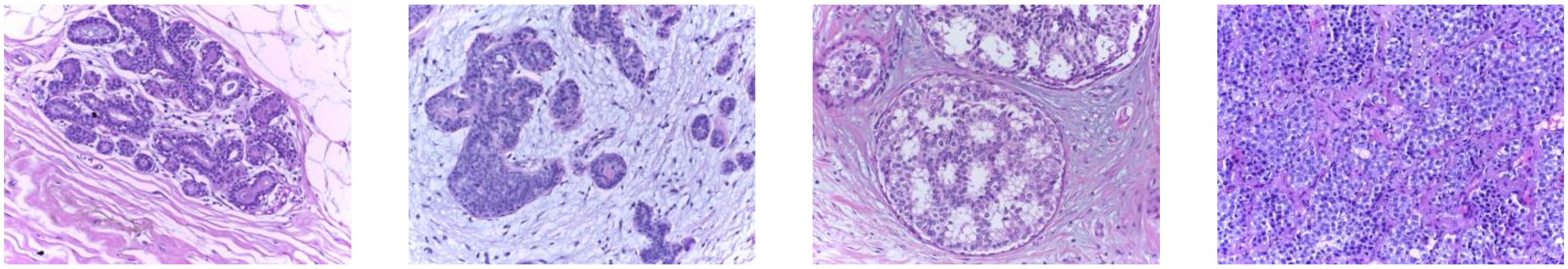

(a) (b) (c) (d)

Fig. 8. Examples of Breast Cancer Histology dataset (BACH). (a) Normal slice, (b) Benign slice, (c) Carcinoma in situ, (d) Invasive carcinoma slice.

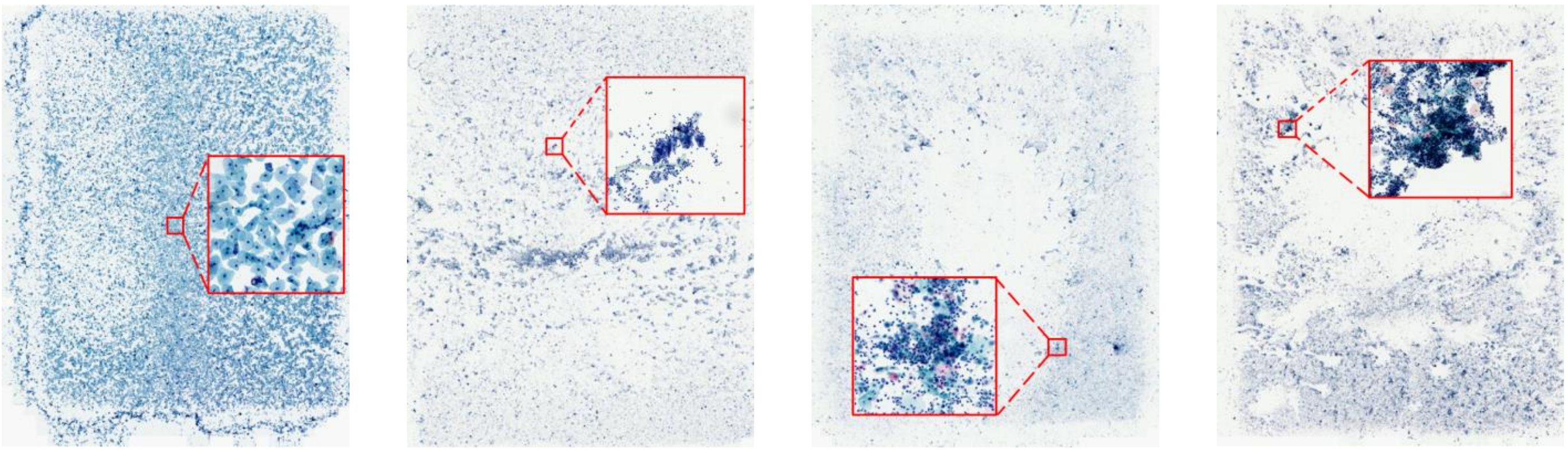

(a) (b) (c) (d)

Fig. 9. Examples of Cervical Carcinoma Grade dataset (CCG). (a) Normal slice, (b) Cancer-level I slice, (c) Cancer-level II slice, (d) Cancer-level III slice. The resolution of slices are gigapixels, i.e. 16473 x 21163. The areas in red squares have been enlarged for illustration.

We notice that some work has already used the stacking atrous convolutions for semantic segmentation [20]. The proposed ADC addresses two primary drawbacks of the existing framework: First, compared to that of the proposed ADC block, the dilation rates used in the existing framework are much larger, i.e. 2, 4, 8, 16. As a result, the receptive field of existing networks normally exceeds the patch size and require lots of padding zeros for convolution computation. Second, the architecture of existing framework has no short-cut connections, which means it is unable to perform multi-scale feature extraction.

#### 2) *Wider Densely Connected Layer*

As the sizes of pathological image datasets are usually small, it is difficult to use them to train an ultra-deep network like the original DenseNet. Ref. [21] demonstrate that the wider network may provide better performance than deeper network on small datasets. Hence, the proposed ADC increases the growth rate ($k$) from 4 (Fig. 6) to 8, 16 and 32, and decreases the number of layers ($l$) from 121 to 28. The proposed dense block is thus wide-and-shallow. To reduce the computation complexity and enhance the capacity of feature representation, the growth rate, i.e. the numbers in ADC modules in Fig. 5, increases as the network goes deeper.

### D. *Implementation*

The proposed ADN is established using *Keras* toolbox. The network was trained with a mini-batch size of 16 on four GPUs (GeForce GTX TITAN X, 12GB RAM). Due to the use of batch normalization layers the initial learning rate was set to a large value, i.e. 0.05, for a faster network convergence. Following that, the learning rate was decreased to 0.01 and then further decreased with gamma = 0.1. The voting approach is adopted to fuse the patch-level predictions made by AND to yield the slice-level class label.

## V. Experimental Results

### A. *Datasets*

#### 1) *Breast Cancer Histology dataset (BACH)*

The BACH dataset [22] consists of 400 Hematoxylin and Eosin (H&E) stained breast histology microscopy images with size 2048 x 1536 pixels, which can be separated to four categories, i.e. normal, benign, in situ carcinoma and invasive carcinoma. Each class has 100 images. The dataset is randomly separated according to the ratio of 80:20 for training and validation. Examples of slices of different categories are shown in Fig. 8. The extra 20 H&E stained breast histological images from Bioimaging dataset [23] are adopted as testing set for the performance comparison between our framework and benchmarking algorithms.

We slide the window with 50% overlapping over the whole image to crop patches with size 512 x 512. The cropping generates 2800 patches for each class. Rotation and mirror were used to increase the size of training dataset. Each patch was rotated by 90°, 180° and 270° and then reflected vertically, resulting in an augmented training set with 896, 00 images. The image-level labels were assigned to the corresponding generated patches.

#### 2) *Cervical Carcinoma Grade dataset (CCG)*

The CCG dataset contains 20 H&E stained whole slice thinprep cytological test (TCT) images, which can be classified to four grades, i.e. normal and cancer-level I, II, III. Each category has five slices, which are separated according to the ratio of 60:20:20 for training, validation and testing. The resolution of TCT slices is 16473 x 21163. Fig. 9 presents the examples of slices of different categories. The CCG dataset was collected by the pathologists collaborating on this project using the whole-slice scanning machine.

TABLE I
DETAILED INFORMATION OF CCG DATASET

| Category | | Training set | Validation set |
|---|---|---|---|
| **Normal** | | 76,576 | 5,676 |
| **Cancer** | **Level I** | 115,164 | 4,105 |
| | **Level II** | 83,712 | 5,336 |
| | **Level III** | 87,380 | 4,742 |

The gigapixel TCT image is too large for deep learning network to directly process. Hence, patches are cropped from the slice images to form the training sets. For each normal slice, around 20,000 224 x 224 patches are randomly cropped. For the cancer slices, shown in Fig. 9, as they have large background areas, we first binarize the TCT slices to detect the region of interest (RoI). Then, the cropping window is slid over the segmented region for patch generation. The image-level label is assigned to the corresponding patches. Rotation was used to increase the size of training dataset. Each patch was rotated by 90°, 180° and 270° to generate an augmented training set with 362,832 images. The patch-level validation set consists of 19,859 patches cropped from the validation slices and verified by the pathologists. The detailed information of patch-level CCG dataset is listed in Table I.

## B. *Evaluation Criterion*

The overall correct classification rate (*ACA*) of all the testing images is adopted as the criterion for performance evaluation. In this section, we will first evaluate the performance of RAL and ADN on the validation set of BACH and CCG. After that, the results of different frameworks on the separate testing sets will be presented.

## C. *Evaluation of RAL*

### 1) *Classification Accuracy during RAL*

The proposed RAL adopts RefineNet (RN) to remove mislabeled patches from training set. As presented in Table II, the size of training set decreases from 89,600 to 86,858 for BACH and from 362,832 to 360,563 for CCG, respectively. Fig. 11 shows some examples of mislabeled patches identified by the RAL, i.e. most of them are normal patches labeled as breast or cervical cancer. The proposed RAL continues until the validation performance stops increasing. The ACAs of patch filtering process on validation set are presented in Table II. It can be observed that the proposed RAL significantly increases the patch-level ACAs of RN, i.e. the improvements for BACH and CCG are 3.65% and 6.01%, respectively.

The t-SNE [24] is used to evaluate the capacity of feature representation of RefineNet during different iterations of BACH training process, as shown in Fig. 10. The points in purple, blue, green and yellow represent the samples of Normal, Benign, Carcinoma in situ and Invasive carcinoma, respectively. It can be observed that the feature representation capacity of the RefineNet gradually improved, i.e. the groups of samples belonged to different categories are separated, during the RAL training on BACH. However, Fig. 10 (e) illustrates that the RefineNet trained after the 4$^{th}$ iteration (K=4) misclassifies some Carcinoma in situ samples (green) to Invasive carcinoma (yellow) and some Normal samples (purple) to Carcinoma in situ samples (green).

### 2) *CNN Models trained with the Refined Dataset*

The RAL refines the training set by removing the mislabeled patches. The information contained in the remaining patches is more accurate and discriminative, which is beneficial for the training of CNN with deeper architecture. To demonstrate the advantage produced by proposed RAL, several well-known deep learning networks like AlexNet [1], VGG-16 [9], ResNet-50/101 [11], DenseNet-121 [12], were involved for the performance evaluation. These networks were trained on the original and refined training set, respectively, and evaluated on the same fully-annotated validation set for both BACH and CCG. The proposed arous DenseNet (AND) is also involved for comparison. The evaluation results are shown in Table III.

TABLE II
PATCH-LEVEL ACA OF RN ON VALIDATION SETS DURING DIFFERENT ITERATIONS OF RAL (%).

| **RAL (**Iteration number K) | **BACH** | | **CCG** | |
|---|---|---|---|---|
| | Training set | Patch-level ACA | Training set | Patch-level ACA |
| trained with original training set (K=0) | 89,600 | 89.16 | 362,832 | 77.87 |
| K=1 | 89,026 | 89.58 | 361,007 | **83.88** |
| K=2 | 88,170 | 89.71 | 360,563 | 82.88 |
| K=3 | 87,363 | **92.81** | - | - |
| K=4 | 86,858 | 92.14 | - | - |

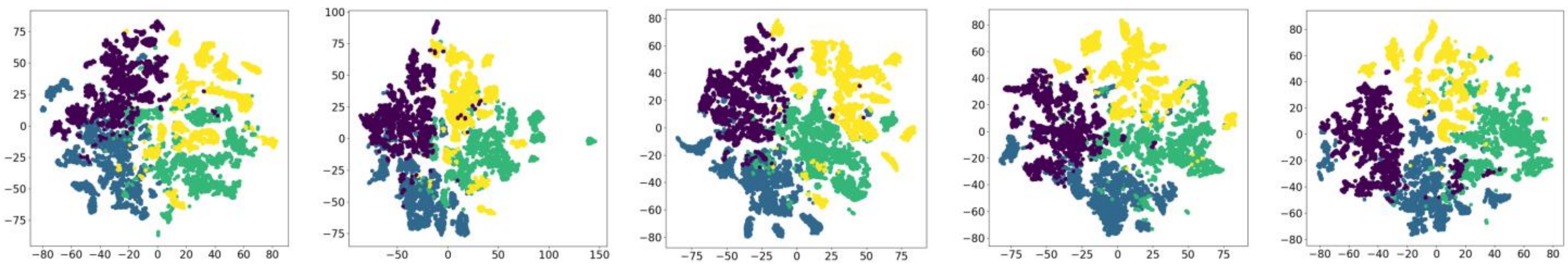

(a) (b) (c) (d) (e)

Fig. 10. The t-SNE figures of the last fully-connected layer of RefineNet for different iteration K of BACH training process. (a)-(e) are for K = 0, 1, 2, 3, 4, respectively.

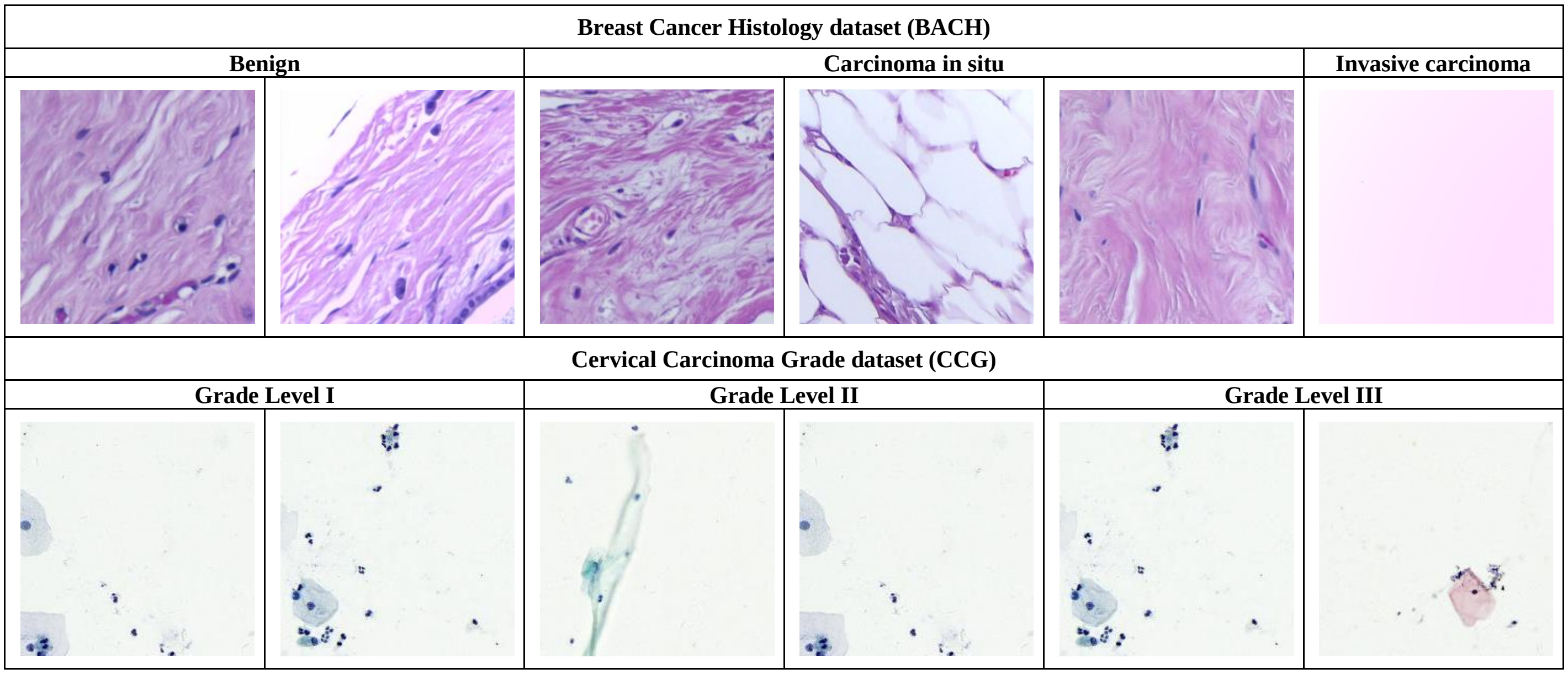

Fig. 11. Illustration of mislabeled patches. The first row lists the normal patches mislabeled to the cancer categories from BACH dataset. The second row lists the mislabeled normal patches from CCG dataset. All the patches have been verified by pathologists.

TABLE III
PATCH-LEVEL AND SLICE-LEVEL VALIDATION ACA OF CNN MODELS TRAINED ON THE ORIGINAL/REFINED TRAINING SETS (%)

| | **BACH** | | | | **CCG** | | | |
|---|---|---|---|---|---|---|---|---|
| | *original* | | *refined* | | *original* | | *refined* | |
| | patch-level | slice-level | patch-level | slice-level | patch-level | slice-level | patch-level | slice-level |
| **AlexNet** [1] | 86.28 | 86.25 | 90.77 | 91.25 | 75.67 | 50 | 82.24 | 75 |
| **VGG-16** [9] | 90.83 | 87.50 | 91.79 | 96.25 | 84.63 | 75 | 90.02 | 75 |
| **ResNet-50** [11] | 89.65 | 86.25 | 92.17 | 93.75 | 79.88 | 75 | 82.31 | 75 |
| **ResNet-101** [11] | 89.05 | 86.25 | 91.17 | 91.25 | 80.06 | 75 | 83.47 | 75 |
| **DenseNet** [12] | 90.39 | 86.25 | 93.29 | 96.25 | 77.87 | 50 | 84.41 | 75 |
| **ADN (ours)** | **91.93** | **88.75** | **94.10** | **97.50** | **85.48** | **75** | **92.05** | **100** |

As shown in Table III, the performances of networks trained on the refined training set are significantly better than that of models trained on the original training set for both BACH and CCG. The highest improvement of patch-level ACA produced by RAL is 4.49% for AlexNet on BACH, and 6.57% for both the AlexNet and our ADN on CCG. For the slice-level ACA, the proposed RAL improves the performance of our ADN from 88.57% to 97.50% on BACH, and from 75% to 100% on CCG, respectively. The results demonstrate that the mislabeled patches in original training set have negative effect for the training of deep learning networks and decrease the classification performances. Furthermore, the refined training set produced by the proposed RAL is useful for general deep learning networks, i.e. shallow network (AlexNet), wide network (VGG-16), multi-branch deep network (ResNet-50) and ultra-deep network (ResNet-101 and DenseNet-121).

## D. *Evaluation of Atrous DenseNet (ADN)*

It can be observed from Table III that our ADN outperforms all the listed benchmarking algorithms on both BACH and CCG without/with the RAL. In this section, a more comprehensive performance analysis of proposed ADN will be presented.

### 1) *ACA on BACH dataset*

The patch-level ACA of different CNN models for each category of BACH is listed in Table IV. All the models were trained with the training set refined by RAL. The average ACA i.e. Ave. ACA, is the overall classification accuracy of the complete patch-level validation set.

TABLE IV
PATCH-LEVEL ACA FOR DIFFERENT CATEGORIES OF BACH (%)

| | **Normal** | **Benign** | **Carcinoma in situ** | **Invasive carcinoma** | **Ave. ACA** |
|---|---|---|---|---|---|
| **AlexNet** [1] | 92.13 | 90.18 | 89.52 | 91.25 | 90.77 |
| **VGG-16** [9] | 90.96 | 93.84 | 89.46 | 92.89 | 91.79 |
| **ResNet-50** [11] | 92.29 | **94.50** | 92.29 | 91.61 | 92.17 |
| **ResNet-101** [11] | 91.96 | 89.20 | 90.66 | 92.88 | 91.17 |
| **DenseNet** [12] | 94.61 | 91.50 | **95.73** | 93.82 | 93.29 |
| **ADN (ours)** | **96.30** | 92.36 | 93.50 | **94.23** | **94.10** |

As shown in Table IV, the proposed ADN achieves the best classification performance for the Normal (96.30%) and Invasive carcinoma (94.23%) patches, while the ResNet-50 and DenseNet-121 yield the highest ACAs for Benign (94.50%) and Carcinoma in situ (95.73%), respectively. The ACAs of our ADN for Benign and Carcinoma in situ are 92.36% and 93.50%, respectively, which are competitive compared to that of the best-performance. The average ACA of ADN is 94.10%, which outperforms the listed benchmarking networks.

To further evaluate the performance of proposed ADN, its confusion map on BACH validation set is presented in Fig. 12. The labels from 1 to 4 stand for Normal, Benign, Carcinoma in situ and Invasive carcinoma, respectively. Fig. 12 illustrates the excellent capacity of proposed ADN for classifying the breast cancer patches.

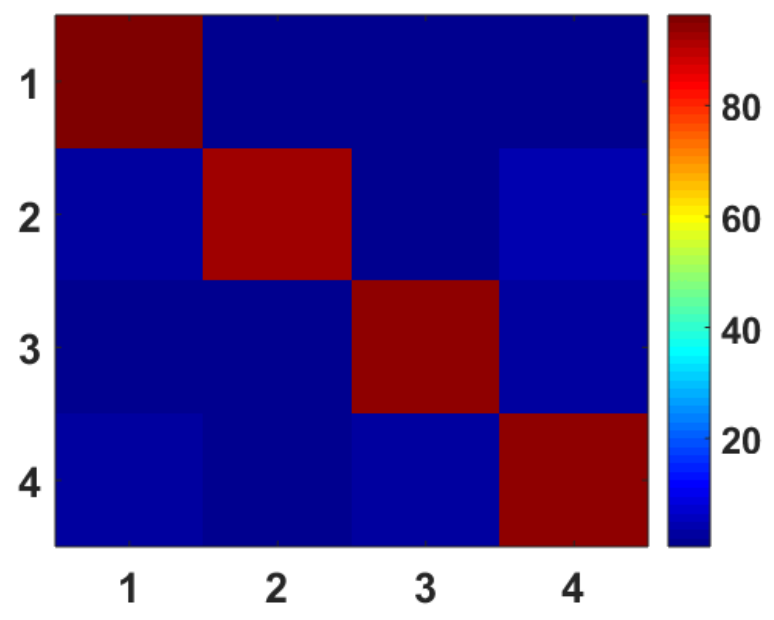

Fig. 12. Confusion map of ADN on BACH validation set

*2) ACA on CCG dataset*

The performance evaluation was also conducted on CCG validation set. Table V listed the experimental results. For the patches from Normal and Level III slices, the proposed ADN achieves the best classification performances, i.e. 99.18% and 70.68%, which are 0.47% and 2.03% higher than that of the first runner-up, i.e. VGG-16. The best ACAs for Level I and Level III patches were achieved by ResNet-50 (99.10%) and ResNet-101 (99.88%), respectively. The proposed ADN yields competitive results for the two categories, which are 97.70% and 99.52%, respectively.

TABLE V
PATCH-LEVEL ACA FOR DIFFERENT CATEGORIES ON CCG (%)

| | **Normal** | **Level I** | **Level II** | **Level III** | **Ave. ACA** |
|---|---|---|---|---|---|
| **AlexNet** [1] | 95.16 | 93.68 | 95.82 | 42.43 | 82.24 |
| **VGG-16** [9] | 98.71 | 96.36 | 98.06 | 65.61 | 90.02 |
| **ResNet-50** [11] | 87.54 | **99.10** | 92.87 | 50.32 | 82.31 |
| **ResNet-101** [11] | 85.46 | 98.32 | **99.88** | 50.45 | 83.47 |
| **DenseNet** [12] | 92.04 | 98.05 | 96.97 | 50.08 | 84.41 |
| **ADN (ours)** | **99.18** | 97.70 | 99.52 | **70.68** | **92.05** |

All the listed algorithms yield relatively low accuracy for the patches from Level III slices. To analyze the reason for the low accuracy, the confusion map of proposed ADN is drawn in Fig. 13. The labels from 1 to 4 stand for Normal, Grade Level I, Grade Level II and Grade Level IIII, respectively. It can be observed that some patches of Grade Level III are wrongly classified to the Normal category. The possible reason is that the lesion area in Grade Level III is smaller than that of Grade Level I and II, which makes the patches cropped from Grade Level III slices usually contain Normal areas. Therefore, the testing Level III patches with large portion of normal area may be recognized as Normal patches by the ADN.

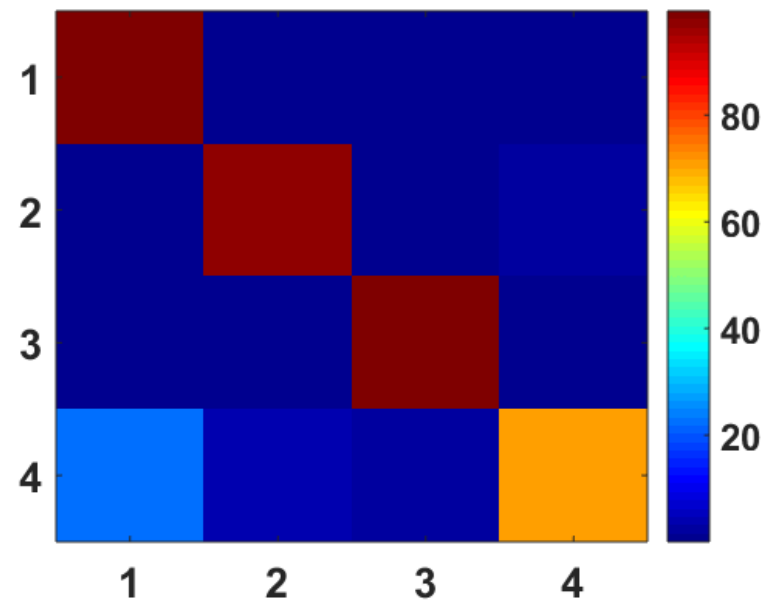

Fig. 13. Confusion map of ADN on CCG validation set

We also evaluated the other deep learning networks and find that they wrongly classify the Level III patches as Normal, either. To address the problem, a suitable approach fusing the patch-level predictions for slice-level decision needs to be developed.

*3) Network Size*

To further illustrate the excellent capacity of proposed ADN, a comparison between different network architectures is presented in Table VI. As aforementioned, instead of building a deeper network, the proposed ADN adopts wider layers to increase its capacity of feature representation, which is more suitable for small dataset. In the experiments, the wider networks, e.g. VGG-16 (16 layers) and ADN (28 layers), achieved better performances than the ultra-deep networks, e.g. ResNet-50/101 (50/101 layers) and DenseNet (121 layers). Since the VGG-16 and ADN have much smaller model size than that of ultra-deep networks, they require fewer network parameters and have lower risk of overfitting to small dataset.

Compared to the straightforward VGG-16, the proposed ADN use multiple atrous convolutions to extract multi-scale features. As shown in Table IV and Table V, the proposed ADN outperforms the VGG-16 and produces the best average ACAs for both BACH (94.10%) and CCG (92.05%) datasets. The experimental results also demonstrate that the proposed ADN can maintain a good balance between network size and capacity of feature learning, which is extremely effective for small pathological datasets.

TABLE VI
DETAILED INFORMATION OF DIFFERENT NETWORK ARCHITECTURES

| | **No. of Layers** | **Model Size** |
|---|---|---|
| **AlexNet** [1] | 8 | 54 M |
| **VGG-16** [9] | 16 | 158 M |
| **ResNet-50** [11] | 50 | 270 M |
| **ResNet-101** [11] | 101 | 488 M |
| **DenseNet** [12] | 121 | 539 M |
| **ADN (ours)** | 28 | 132 M |

## *E. Results on the Testing Set*

In this section, we compare the performances of different frameworks on the testing set of BACH and CCG.

TABLE VII
ACA OF DIFFERENT FRAMEWORKS FOR BACH TESTING SET (%)

| | Patch-level | | | | | Slice-level |
|---|---|---|---|---|---|---|
| | **Normal** | **Benign** | **Carcinoma in situ** | **Invasive carcinoma** | **Ave. ACA** | **Ave. ACA** |
| **CNN** [23] | 61.70 | 56.70 | 83.30 | 88.30 | 72.50 | 80.00 |
| **CNN+SVM** [23] | 65.00 | 61.70 | 76.70 | 88.30 | 72.93 | 85.00 |
| **AlexNet** [1] | 60.00 | 58.33 | 85.00 | 95.00 | 74.58 | 80.00 |
| **VGG-16** [9] | 75.00 | 61.67 | 75.00 | 90.00 | 75.42 | 85.00 |
| **ResNet-50** [11] | 63.33 | 65.00 | 80.00 | 95.00 | 75.83 | 85.00 |
| **ResNet-101** [11] | 65.00 | 70.00 | 75.00 | 90.00 | 75.00 | 85.00 |
| **DenseNet** [12] | 66.67 | **76.67** | 73.33 | 88.33 | 76.25 | 85.00 |
| **ADN (ours)** | 60.00 | 66.67 | 88.33 | 93.33 | 77.08 | 85.00 |
| **ADN + RAL (ours)** | **71.67** | 73.33 | **88.33** | **96.67** | **82.50** | **90.00** |

TABLE VIII
ACA OF DIFFERENT FRAMEWORKS FOR CCG TESTING SET (%)

| | Patch-level | | | | | Slice-level |
|---|---|---|---|---|---|---|
| | **Normal** | **Level I** | **Level II** | **Level III** | **Ave. ACA** | **Ave. ACA** |
| **AlexNet** [1] | 91.75 | 42.24 | 69.88 | 70.91 | 68.70 | 50 |
| **VGG-16** [9] | 97.80 | 63.65 | 71.25 | 78.39 | 77.77 | 75 |
| **ResNet-50** [11] | 97.82 | 46.86 | 75.05 | 68.57 | 72.08 | 50 |
| **ResNet-101** [11] | 96.64 | 67.34 | 75.57 | 58.66 | 74.55 | 50 |
| **DenseNet** [12] | 98.81 | 56.62 | 72.20 | 71.04 | 74.67 | 75 |
| **ADN (ours)** | 99.29 | 71.51 | 76.51 | 73.81 | 80.28 | 75 |
| **ADN + RAL (ours)** | **99.95** | **80.35** | **85.31** | **82.60** | **87.05** | **100** |

#### 1) *Patch-level and Slice-level ACA on BACH*

The extra 20 H&E stained breast histological images from public available dataset, i.e. Bioimaging [23], are employed as testing set to evaluate the performances of frameworks trained on BACH dataset. The results on the testing set are listed in Table VII.

As shown in Table VII, the proposed ADN achieves the best average patch-level classification performances, i.e. 77.08%, on the testing set, which is 0.83% higher than the runner-up, i.e. DenseNet-121. The proposed RAL further leads to an improvement of 5.42% for the final classification accuracy. Accordingly, the slice-level average classification accuracy (90%) of the proposed ADN + RAL framework ranks the best among the listed benchmarking algorithms.

#### 2) *Patch-level and Slice-level ACA on CCG*

The results for the testing set of CCG are listed in Table VIII. It can be observed that the proposed ADN achieved the best patch-level ACA, i.e. 80.28%, among the models trained with the original training set, which is 2.51% higher than the runner-up, i.e. VGG-16. Furthermore, we noticed that most of the listed benchmark algorithms don't do well for the patches of grade level I, i.e. the highest accuracy produced by the ultra-deep ResNet-101is only 67.34%. Due to the excellent network design with a 28-layers architecture, our ADN successfully distinguishes the grade level I patches from others, i.e. a patch-level ACA of 71.51% was achieved. The accuracy is 4.17% higher than that of ResNet-101 (101 layers).

The proposed RAL refines the training set by removing the mislabeled patches, which benefits the following network training. As a result, the RAL training strategy yields significant improvements of both average patch-level ACA (6.77%) and average slice-level ACA (25%) for our ADN framework.

## VI. CONCLUSION

Increasing number of researches adopt deep learning based framework for medical image analysis. However, the usage of deep learning networks for the pathological image analysis encounters several challenges. For example, most of pathological images have high resolution, i.e. gigapixel. It is difficult for CNN to directly process the gigapixel images, due to the expensive computational cost. Cropping patches from the whole slice images is the common approach to address the problem. However, most of the pathological datasets only have the image-level labels. While the image level labels can be correspondingly assigned to the cropped patches, the patch-level training set usually contains mislabeled samples.

To address the challenges, we proposed a complete framework for the pathological image classification. The framework consists of a novel training strategy, namely reversed active learning (RAL), and an advanced network architecture, namely atrous DenseNet (ADN). The proposed reversed active learning (RAL) can remove the mislabel patches in the training set. The refined training set can then be used to train other widely used deep learning networks, e.g. VGG-16, ResNets, etc. A novel deep learning network, i.e. atrous DenseNet (ADN), is also proposed for the classification of pathological images. The proposed ADN achieves multi-scale feature extraction by combining the atrous convolutions and Dense Block.

The proposed RAL and ADN have been evaluated on two pathological datasets, i.e. BACH and CCG. The experimental results demonstrate the excellent performance of the proposed ADN + RAL framework, i.e. average patch-level ACAs of 94.10% and 92.05% on BACH and CCG validation sets were achieved.

## References

[1] Krizhevsky, A., I. Sutskever, and G.E. Hinton, *ImageNet classification with deep convolutional neural networks*, in *International Conference on Neural Information Processing Systems*. 2012. p. 1097-1105.

[2] Liu, Y., K. Gadepalli, M. Norouzi, G.E. Dahl, T. Kohlberger, A. Boyko, S. Venugopalan, A. Timofeev, P.Q. Nelson, and G.S. Corrado, *Detecting cancer metastases on gigapixel pathology images.* arXiv e-print arXiv:1703.02442, 2017.

[3] Hou, L., D. Samaras, T.M. Kurc, Y. Gao, J.E. Davis, and J.H. Saltz. *Patch-based convolutional neural network for whole slide tissue image classification*. in *IEEE Conference on Computer Vision and Pattern Recognition*. 2016.

[4] Wang, D. and Y. Shang. *A new active labeling method for deep learning*. in *International Joint Conference on Neural Networks*. 2014.

[5] Rahhal, M.M.A., Y. Bazi, H. Alhichri, N. Alajlan, F. Melgani, and R.R. Yager, *Deep learning approach for active classification of electrocardiogram signals.* Information Sciences, 2016. **345**(C): p. 340-354.

[6] Yang, L., Y. Zhang, J. Chen, S. Zhang, and D.Z. Chen, *Suggestive annotation: a deep active learning framework for biomedical image segmentation*, in *International Conference on Medical Image Computing and Computer-Assisted Intervention*. 2017. p. 399-407.

[7] Zhou, Z., J. Shin, L. Zhang, S. Gurudu, M. Gotway, and J. Liang. *Fine-Tuning Convolutional Neural Networks for Biomedical Image Analysis: Actively and Incrementally*. in *IEEE Conference on Computer Vision and Pattern Recognition*. 2017.

[8] Nan, Y., G. Coppola, Q. Liang, K. Zou, W. Sun, D. Zhang, Y. Wang, and G. Yu, *Partial labeled gastric tumor segmentation via patch-based reiterative learning.* arXiv e-print arXiv:1712.07488, 2017.

[9] Simonyan, K. and A. Zisserman, *Very deep convolutional networks for large-scale image recognition.* arXiv e-print arXiv:1409.1556, 2015.

[10] Szegedy, C., W. Liu, Y. Jia, and P. Sermanet, *Going deeper with convolutions*, in *IEEE Conference on Computer Vision and Pattern Recognition*. 2015. p. 1-9.

[11] He, K., X. Zhang, S. Ren, and J. Sun, *Deep residual learning for image recognition*, in *IEEE Conference on Computer Vision and Pattern Recognition*. 2016. p. 770-778.

[12] Huang, G., Z. Liu, L.V.D. Maaten, and K.Q. Weinberger, *Densely Connected Convolutional Networks*, in *IEEE Conference on Computer Vision and Pattern Recognition*. 2017. p. 2261-2269.

[13] Albarqouni, S., C. Baur, F. Achilles, V. Belagiannis, S. Demirci, and N. Navab, *AggNet: deep learning from crowds for mitosis detection in breast cancer histology images.* IEEE Transactions on Medical Imaging, 2016. **35**(5): p. 1313-1321.

[14] Shah, M., C. Rubadue, D. Suster, and D. Wang, *Deep Learning Assessment of Tumor Proliferation in Breast Cancer Histological Images.* arXiv e-print arXiv:1610.03467, 2016.

[15] Chen, H., X. Qi, L. Yu, and P.A. Heng, *DCAN: deep contour-aware networks for accurate gland segmentation*, in *IEEE Conference on Computer Vision and Pattern Recognition*. 2016. p. 2487-2496.

[16] Li, Y., L. Shen, and S. Yu, *HEp-2 Specimen Image Segmentation and Classification Using Very Deep Fully Convolutional Network.* IEEE Transactions on Medical Imaging, 2017. **36**(7): p. 1561-1572.

[17] Liu, J., B. Xu, C. Zheng, Y. Gong, J. Garibaldi, D. Soria, A. Green, I.O. Ellis, W. Zou, and G. Qiu, *An End-to-End Deep Learning Histochemical Scoring System for Breast Cancer Tissue Microarray.* arXiv e-print arXiv:1801.06288, 2018.

[18] Yu, F. and V. Koltun, *Multi-scale context aggregation by dilated convolutions.* arXiv e-print arXiv:1511.07122, 2016.

[19] Li, Y. and L. Shen, *HEp-Net: a smaller and better deep-learning network for HEp-2 cell classification.* Computer Methods in Biomechanics and Biomedical Engineering: Imaging & Visualization, 2018: p. 1-7.

[20] Chen, L.C., G. Papandreou, F. Schroff, and H. Adam, *Rethinking atrous convolution for semantic image segmentation.* arXiv e-print arXiv:1706.05587, 2017.

[21] Zagoruyko, S. and N. Komodakis, *Wide Residual Networks.* arXiv e-print arXiv:1605.07146, 2016.

[22] Araújo, T., G. Aresta, C. Eloy, A. Polónia, and P. Aguiar. *ICIAR 2018 Grand Challenge on BreAst Cancer Histology images*. 2018; Available from: https://iciar2018-challenge.grand-challenge.org/home/.

[23] Araújo, T., G. Aresta, E. Castro, J. Rouco, P. Aguiar, C. Eloy, A. Polónia, and A. Campilho, *Classification of breast cancer histology images using Convolutional Neural Networks.* Plos One, 2017. **12**(6): p. e0177544.

[24] van der Maaten, L. and G. Hinton, *Visualizing high-dimensional data using t-SNE.* J. Mach. Learn. Res., 2008. **9**: p. 2579-2605.

## Appendix

To alleviate the overfitting problem, a simple CNN, namely RefineNet (RN), was adopted in the iterative Reverse Active Learning (RAL) process to remove mislabeled patches. The pipeline of RefineNet is presented in Table IX, which consists of convolutional (C), max pooling (MP), averaging pooling (AP) and fully-connected (FC) layers.

Table IX
Architecture of RN. Pipeline consists of Convolutional layer (C), Max pooling layer (MP), Average pooling layer (AP) and Fully-connected layer (FC).

| **Layer** | **Type** | **Kernel size & number** |
|---|---|---|
| 1 | C | 3x3, 16 |
| 2 | MP | 2x2 |
| 3 | C | 3x3, 32 |
| 4 | MP | 2x2 |
| 5 | C | 3x3, 64 |
| 6 | MP | 2x2 |
| 7 | C | 3x3, 64 |
| 8 | MP | 2x2 |
| 9 | C | 3x3, 128 |
| 10 | MP | 2x2 |
| 11 | C | 3x3, 128 |
| 12 | AP | 7x7 |
| 13 | FC | 256 |
| 14 | FC | 4 |